\pdfoutput=1
\documentclass[11pt,a4paper]{article}
\usepackage{authblk}
\usepackage[hyperref]{eacl2021}
\usepackage{times}
\usepackage{latexsym}
\usepackage{longtable}

\usepackage{microtype}
\usepackage{breakurl}

\usepackage{booktabs}
\usepackage{graphicx}
\usepackage[T1]{fontenc}
\usepackage{textcomp}
\newcommand*\rot{\rotatebox{90}}

\aclfinalcopy % Uncomment this line for the final submission
\def\aclpaperid{1} %  Enter the acl Paper ID here

\title{HerBERT: Efficiently Pretrained Transformer-based \\ Language Model for Polish}

\author[1]{Robert Mroczkowski}
\author[1]{Piotr Rybak}
\author[2]{Alina Wróblewska}
\author[1,3]{Ireneusz Gawlik}
\affil[1]{ML Research at Allegro.pl}
\affil[2]{Institute of Computer Science, Polish Academy of Sciences} 
\affil[3]{Department of Computer Science, AGH University of Science and Technology} 
\affil[ ]{\texttt{\{firstname.lastname\}@allegro.pl, alina@ipipan.waw.pl}}

\date{}

\begin{document}
\maketitle
\begin{abstract}
BERT-based models are currently used for solving nearly all Natural Language Processing (NLP) tasks and most often achieve state-of-the-art results. Therefore, the NLP community conducts extensive research on understanding these models, but above all on designing effective and efficient training procedures. Several ablation studies investigating how to train BERT-like models have been carried out, but the vast majority of them concerned only the English language. A training procedure designed for English does not have to be universal and applicable to other especially typologically different languages. Therefore, this paper presents the first ablation study focused on Polish, which, unlike the isolating English language, is a fusional language. We design and thoroughly evaluate a pretraining procedure of transferring knowledge from multilingual to monolingual BERT-based models. In addition to multilingual model initialization, other factors that possibly influence pretraining are also explored, i.e. training objective, corpus size, BPE-Dropout, and pretraining length. Based on the proposed procedure, a Polish BERT-based language model -- HerBERT -- is trained. This model achieves state-of-the-art results on multiple downstream tasks.
\end{abstract}

\section{Introduction}
\label{sec:intro}
Recent advancements in self-supervised pretraining techniques drastically changed the way we design Natural Language Processing (NLP) systems. Even though, pretraining has been present in NLP for many years \citep{mikolov2013distributed, pennington-etal-2014-glove, bojanowski-etal-2017-enriching}, only recently we observed a shift from task-specific to general-purpose models. In particular, the BERT model \citep{devlin2019bert} proved to be a dominant architecture and obtained state-of-the-art results for a variety of NLP tasks.

While most of the research related to analyzing and improving BERT-based models focuses on English, there is an increasing body of work aimed at training and evaluation of models for other languages, including Polish. Thus far, a handful of models specific for Polish has been released, e.g. Polbert\footnote{\url{https://github.com/kldarek/polbert}}, first version of HerBERT \citep{rybak-etal-2020-klej}, and Polish RoBERTa \citep{dadas2020pretraining}.

Aforementioned works lack ablation studies, making it difficult to attribute hyperparameters choices to models performance. In this work, we fill this gap by conducting an extensive set of experiments and developing an efficient BERT training procedure. As a result, we were able to train and release a new BERT-based model for Polish language understanding. Our model establishes a new state-of-the-art on the variety of downstream tasks including semantic relatedness, question answering, sentiment analysis and part-of-speech tagging.

To summarize, our contributions are:
\begin{enumerate}
    \item development and evaluation of an efficient pretraining procedure for transferring knowledge from multilingual to monolingual language models based on work by \citet{arkhipov-etal-2019-slavic},
    \item detailed analysis and an ablation study challenging the effectiveness of Sentence Structural Objective (SSO,  \citealp{structbert2020iclr}), and Byte Pair Encoding Dropout (BPE-Dropout,  \citealp{bpedropout2020provilkov}),
    \item release of HerBERT\footnote{\url{https://huggingface.co/allegro/herbert-large-cased}} -- a BERT-based model for Polish language understanding, which achieves state-of-the-art results on KLEJ Benchmark \citep{rybak-etal-2020-klej} and POS tagging task \citep{wroblewska-2020-towards}.
\end{enumerate}

The rest of the paper is organized as follows. In Section \ref{sec:related}, we provide an overview of related work. After that, Section \ref{sec:language_model} introduces the BERT-based language model and experimental setup used in this work. In Section \ref{sec:ablation}, we conduct a thorough ablation study to investigate the impact of several design choices on the performance of downstream tasks. Next, in Section \ref{sec:herbert} we apply drawn conclusions and describe the training of HerBERT model. In Section \ref{sec:eval}, we evaluate HerBERT on a set of eleven tasks and compare its performance to other state-of-the-art models. Finally, we conclude our work in Section \ref{sec:conclusion}.

\section{Related Work}
\label{sec:related}
The first significant ablation study of BERT-based language pretraining was described by \citet{liu2019roberta}. Authors demonstrated the ineffectiveness of Next Sentence Prediction (NSP) objective, the importance of dynamic token masking, and gains from using both large batch size and large training dataset. Further large-scale studies analyzed the relation between the model and the training dataset sizes \citep{kaplan2020scaling}, the amount of compute used for training \citep{DBLP:journals/corr/abs-2005-14165-gpt3} and training strategies and objectives \citep{DBLP:journals/corr/abs-1910-10683-T5}.

Other work focused on studying and improving BERT training objectives. As mentioned before, the NSP objective was either removed \citep{liu2019roberta} or enhanced either by predicting the correct order of sentences (Sentence Order Prediction (SOP), \citealp{albert2020iclr}) or discriminating between previous, next and random sentence (Sentence Structural Objective (SSO), \citealp{structbert2020iclr}). Similarly, the Masked Language Modelling (MLM) objective was extended to either predict spans of tokens \citep{joshi2019spanbert}, re-order shuffled tokens (Word Structural Objective (WSO),  \citealp{structbert2020iclr}) or replaced altogether with a binary classification problem using mask generation \citep{DBLP:conf/iclr/ClarkLLM20-electra}.

For tokenization, the Byte Pair Encoding algorithm (BPE, \citealp{sennrich2016bpe}) is commonly used. The original BERT model used WordPiece implementation \citep{schuster2012japanese}, which was later replaced by SentencePiece \citep{kudo-richardson-2018-sentencepiece}. \citet{NIPS2018_7408-frage} discovered that rare words lack semantic meaning. \citet{bpedropout2020provilkov} proposed a BPE-Dropout technique to solve this issue.

All of the above work was conducted for English language understanding. There was little research into understanding how different pretraining techniques affect BERT-based models for other languages. The main research focus was to train BERT-based models and report their performance on downstream tasks. The first such models were released for German\footnote{\url{https://deepset.ai/german-bert}} and Chinese \citep{devlin2019bert}, recently followed by Finnish \citep{virtanen2019multilingual}, 
French \citep{martin-etal-2020-camembert,le-etal-2020-flaubert}, 
Polish \citep{rybak-etal-2020-klej, dadas2020pretraining}, 
Russian \citep{kuratov2019adaptation}, 
and many other languages\footnote{\url{https://huggingface.co/models}}.
Research on developing and investigating an efficient procedure of pretraining BERT-based models was rather neglected in these languages.

Language understanding for low-resource languages has also been addressed by training jointly for several languages at the same time. That approach improves performance for moderate and low-resource languages as showed by \citet{NIPS2019_8928_XLM}. The first model of this kind was the multilingual BERT trained for 104 languages \citep{devlin2019bert} followed by
\citet{NIPS2019_8928_XLM} and \citet{conneau-etal-2020-xlmr}.

\section{Experimental Setup}
\label{sec:language_model}
In this section, we describe the experimental setup used in the ablation study. First, we introduce the corpora we used to train models. Then, we give an overview of the language model architecture and training procedure. In particular, we describe the method of transferring knowledge from multilingual to monolingual BERT-based models. Finally, we present the evaluation tasks.

\subsection{Training Data}
\label{subsec:data}
We gathered six corpora to create two datasets on which we trained HerBERT. The first dataset (henceforth called \emph{Small}) consists of corpora of the highest quality, i.e. NKJP, Wikipedia, and Wolne Lektury. The second dataset (\emph{Large}) is over five times larger as it additionally contains texts of lower quality (CCNet and Open Subtitles). Below, we present a short description of each corpus. Additionally, we include the basic corpora statistics in Table \ref{tab:corpora}.

\begin{table}[!ht]
\renewcommand*{\arraystretch}{1.15}
\setlength{\tabcolsep}{0.38em}
\centering
\begin{tabular}{lrrr}

    \toprule
    \bf{Corpus} & \bf{Tokens} & \bf{Documents} & \bf{Avg len} \\
    \toprule
    \multicolumn{4}{c}{\bf{Source Corpora}} \\
    \midrule
    NKJP & 1357M & 3.9M & 347 \\
    Wikipedia & 260M & 1.4M & 190 \\
    Wolne Lektury & 41M & 5.5k & 7447 \\
    \midrule
    CCNet Head & 2641M & 7.0M & 379 \\
    CCNet Middle & 3243M & 7.9M & 409 \\
    Open Subtitles & 1056M & 1.1M & 961 \\
    \midrule
    \multicolumn{4}{c}{\bf{Final Corpora}} \\
    \midrule
    Small & 1658M & 5.3M & 313 \\
    Large & 8599M & 21.3M & 404 \\
    \bottomrule

\end{tabular}
\caption{Overview of all data sources used to train HerBERT. We combine them into two corpora. The \emph{Small} corpus consists of the highest quality text resources: NKJP, Wikipedia, and Wolne Lektury. The \emph{Large} corpus consists of all sources. \textit{Avg len} is the average number of tokens per document in each corpus.}
\label{tab:corpora}
\end{table}

\paragraph{NKJP} (Narodowy Korpus Języka Polskiego, eng. \emph{National Corpus of Polish}) \citep{przepiorkowski2012narodowy} is a well balanced collection of Polish texts. It consists of texts from many different sources, such as classic literature, books, newspapers, journals, transcripts of conversations, and texts crawled from the internet.

\paragraph{Wikipedia} is an online encyclopedia created by the community of Internet users. Even though it is crowd-sourced, it is recognized as a high-quality collection of articles.

\paragraph{Wolne Lektury} (eng. \emph{Free Readings})\footnote{\url{https://wolnelektury.pl}} is a collection of over five thousand books and poems, mostly from 19th and 20th century, which have already fallen in the public domain.

\paragraph{CCNet}\citep{wenzek-etal-2020-ccnet} is a clean monolingual corpus extracted from Common Crawl\footnote{\url{http://commoncrawl.org/}} dataset of crawled websites. 

\paragraph{Open Subtitles} is a popular website offering movie and TV subtitles, which was used by \citet{LISON16.947} to curate and release a multilingual parallel corpus from which we extracted its monolingual Polish part. 

\subsection{Language Model}
\label{subsec:model}

\paragraph{Tokenizer}
We used Byte-Pair Encoding (BPE) tokenizer \citep{sennrich2016bpe} with the vocabulary size of 50k tokens and trained it on the most representative parts of our corpus, i.e annotated subset of the NKJP, and the Wikipedia.

Subword regularization is supposed to emphasize the semantic meaning of tokens \citep{NIPS2018_7408-frage,bpedropout2020provilkov}. To verify its impact on training language model we used a BPE-Dropout \citep{bpedropout2020provilkov} with a probability of dropping a merge equal to 10\%.

\paragraph{Architecture}

We followed the original BERT \citep{devlin2019bert} architectures for both BASE (12 layers, 12 attention heads and hidden dimension of 768) and LARGE (24 layers, 16 attention heads and hidden dimension of 1024) variants.

\paragraph{Initialization}
We initialized models either randomly or by using weights from XLM-RoBERTa \citep{conneau-etal-2020-xlmr}. In the latter case, the parameters for all layers except word embeddings and token type embeddings were copied directly from the source model. Since XLM-RoBERTa does not use the NSP objective and does not have the token type embeddings, we took them from the original BERT model.

To overcome the difference in tokenizers vocabularies we used a method similar to \citep{arkhipov-etal-2019-slavic}. If a token from the target model vocabulary was present in the source model vocabulary then we directly copied its weights. 

Otherwise, it was split into smaller units and the embedding was obtained by averaging sub-tokens embeddings.

\paragraph{Training Objectives}
\label{par:training_objective}
We trained all models with an updated version of the MLM objective \citep{joshi2019spanbert, martin-etal-2020-camembert}, masking ranges of subsequent tokens belonging to single words instead of individual (possibly subword) tokens. We replaced the NSP objective with SSO. The other parameters were kept the same as in the original BERT paper. Training objective is defined in Equation \ref{eq:loss}.

\begin{equation}
\label{eq:loss}
\mathcal{L} = \mathcal{L}_{\mathrm{MLM}}(\theta) + \alpha\cdot\mathcal{L}_{\mathrm{SSO}}(\theta)
\end{equation}

where $\alpha$ is the SSO weight.

\subsection{Tasks}
\label{subsec:tasks}

\paragraph{KLEJ Benchmark}
\label{par:klej_benchmark}
The standard method for evaluating pretrained language models is to use a diverse collection of tasks grouped into a single benchmark. Such benchmarks exist in many languages, e.g. English (GLUE, \citealp{wang2019glue}), Chinese (CLUE, \citealp{xu2020clue}), and Polish (KLEJ, \citealp{rybak-etal-2020-klej}). 

Following this paradigm we first verified the quality of assessed models with KLEJ. It consists of nine tasks: name entity classification (\textbf{NKJP-NER}, \citealp{przepiorkowski2012narodowy}), semantic relatedness (\textbf{CDSC-R}, \citealp{wroblewska2017polish}), natural language inference (\textbf{CDSC-E}, \citealp{wroblewska2017polish}), cyberbullying detection (\textbf{CBD}, \citealp{ptaszynski2019results}), sentiment analysis (\textbf{PolEmo2.0-IN}, \textbf{PolEmo2.0-OUT}, \citealp{kocon-etal-2019-multi}, \textbf{AR}, \citealp{rybak-etal-2020-klej}), question answering (\textbf{Czy wiesz?}, \citealp{marcinczuk2013open}), and text similarity (\textbf{PSC}, \citealp{ogro:kop:14:lrec}).

\paragraph{POS Tagging and Dependency Parsing}
All of the KLEJ Benchmark tasks belong to the classification or regression type. It is therefore difficult to assess the quality of individual token embeddings. To address this issue, we further evaluated HerBERT on part-of-speech tagging and dependency parsing tasks.

For tagging, we used the manually annotated subset of NKJP \cite{1mln_nkjp}, converted to the CoNLL-U format by \citet{wroblewska-2020-towards}. 
We evaluated models performance on a test set using accuracy and F1-Score. 

For dependency parsing, we applied Polish Dependency Bank \citep{pl} from the Universal Dependencies repository (release 2.5, \citealp{ud25data}).

In addition to three Transformer-based models, we also included models trained with static embeddings. The first one did not use pretrained embeddings while the latter utilized fastText \citep{bojanowski-etal-2017-enriching} embeddings trained on Common Crawl.

The models are evaluated with the standard metrics: UAS (unlabeled attachment score) and LAS (labelled attachment score). The gold-standard segmentation was preserved. We report the results on the test set. 

\section{Ablation Study}
\label{sec:ablation}
In this section, we analyze the impact of several design choices on downstream task performance of Polish BERT-based models. In particular, we focus on initialization, corpus size, training objective, BPE-Dropout, and the length of pretraining. 

\subsection{Experimental Design}
\paragraph{Hyperparameters}
Unless stated otherwise, in all experiments we trained BERT\textsubscript{BASE} model initialized with XLM-RoBERTa weights for 10k iterations using a linear decay schedule of the learning rate with a peak value of $7 \cdot 10^{-4}$ and a warm-up of 500 iterations. We used a batch size of 2560.

\paragraph{Evaluation}
Different experimental setups were compared using the average score on the KLEJ Benchmark. The validation sets are used for evaluation and we report the results corresponding to the median values of the five runs. Since only six tasks in KLEJ Benchmark have validation sets the scores are not directly comparable to those reported in Section \ref{sec:eval}. We used Welch's t-test \citep{10.1093/biomet/34.1-2.28-welch} with a p-value of 0.01 to test for statistical differences between experimental variants.

\subsection{Results}

\paragraph{Initialization}
\label{subsec:init-study}
One of the main goals of this work is to propose an efficient strategy to train a monolingual language model. We began with investigating the impact of pretraining the language model itself. For this purpose, the following experiments were designed.

\begin{table}[!h]
\renewcommand*{\arraystretch}{1.15}
\centering
\begin{tabular}{l|llr}

    \toprule
    \bf{Init} & \bf{Pretraining} & \bf{BPE} & \bf{Score} \\
    \midrule
    \multicolumn{4}{c}{\bf{Ablation Models}} \\
    \midrule
    Random & No & - & 58.15 $\pm$ 0.33 \\
    XLM-R & No & - & \underline{83.15 $\pm$ 1.22} \\
    \midrule
    Random & Yes & No & 85.65 $\pm$ 0.43 \\
    XLM-R & Yes & No & \underline{88.80 $\pm$ 0.15} \\
    \midrule
    Random & Yes & Yes & 85.78 $\pm$ 0.23 \\
    XLM-R & Yes & Yes & \bf{\underline{89.10 $\pm$ 0.19}} \\
    \midrule
    \multicolumn{4}{c}{\bf{Original Models}} \\
    \midrule
    XLM-R & - & - & 88.82 $\pm$ 0.15 \\
    \bottomrule

\end{tabular}
\caption{Average scores on KLEJ Benchmark depending on the initialization scheme: Random -- initialization with random weights, XLM-R -- initialization with XLM-RoBERTa weights. We used BERT\textsubscript{BASE} model trained for 10k iterations with the SSO weight equal to $1.0$ on the \emph{Large} corpus. The best score within each group is underlined, the best overall is bold.}
\label{tab:init-study}
\end{table}

First, we fine-tuned randomly initialized BERT model on KLEJ Benchmark tasks. Note that this model is not pretrained in any way. As expected, the results on the KLEJ Benchmark are really poor with the average score equal to 58.15. 

Next, we evaluated the BERT model initialized with XLM-RoBERTa weights (see Table \ref{tab:init-study}). It achieved much better average score than the randomly initialized model (83.15 vs 58.15), but it was still not as good as the original XLM-RoBERTa model (88.82\%). The difference in the performance can be explained by the transfer efficiency. The method of transferring token embeddings between different tokenizers proves to retain most information, but not all of it.

To measure the impact of initialization on pretraining optimization, we trained the aforementioned models for 10k iterations. Beside MLM objective, we used SSO loss with $\alpha=1.0$ and conducted experiments with both enabled and disabled BPE-Dropout. Models initialized with XLM-RoBERTa achieve significantly higher results than models initialized randomly, 89.10 vs 85.78 and 88.80 vs 85.65 for pretraining with and without BPE-Dropout respectively.

Models initialized with XLM-RoBERTa achieved similar results to the original XLM-RoBERTa (the differences are not statistically significant). It proves that it is possible to quickly recover from the performance drop caused by a tokenizer conversion procedure and obtain a much better model than the one initialized randomly.

\paragraph{Corpus Size}
\label{subsec:corpus-size}
As mentioned in Section \ref{sec:related}, previous research show that pretraining on a larger corpus is beneficial for downstream task performance \citep{kaplan2020scaling, DBLP:journals/corr/abs-2005-14165-gpt3}. We investigated this by pretraining BERT\textsubscript{BASE} model on both \emph{Small} and \emph{Large} corpora (see Section \ref{subsec:data}). To mitigate a possible impact of confounding variable, we also vary the weight of SSO loss and usage of BPE-Dropout (see Table \ref{tab:corpus-size-study}).

\begin{table}[!ht]
\renewcommand*{\arraystretch}{1.15}
\centering
\begin{tabular}{l|llr}

    \toprule
    \bf{Corpus} & \bf{SSO} & \bf{BPE} & \bf{Score} \\
    \midrule
    Small & 1.0 & Yes & 88.73 $\pm$ 0.08 \\
    Large & 1.0 & Yes & \underline{89.10 $\pm$ 0.19} \\
    \midrule
    Small & 0.1 & Yes & 88.90 $\pm$ 0.24 \\
    Large & 0.1 & Yes & \bf{\underline{89.37 $\pm$ 0.25}} \\
    \midrule
    Small & 0.0 & Yes & 89.18 $\pm$ 0.15 \\
    Large & 0.0 & Yes & \underline{89.25 $\pm$ 0.21} \\
    \midrule
    Small & 0.0 & No & 89.12 $\pm$ 0.29 \\
    Large & 0.0 & No & \underline{89.28 $\pm$ 0.26} \\
    \bottomrule

\end{tabular}
\caption{Average scores on KLEJ Benchmark depending on a corpus size. We used BERT\textsubscript{BASE} model trained for 10k iterations with or without BPE-Dropout and with various SSO weights. The best score within each group is underlined, the best overall is bold.}
\label{tab:corpus-size-study}
\end{table}

As expected, the model pretrained on a \emph{Large} corpus performs better on downstream tasks. However, the difference is statistically significant only for the experiment with SSO weight equal to 1.0 and BPE-Dropout enabled. Therefore it's not obvious whether a larger corpus is actually beneficial.

\paragraph{Sentence Structural Objective}
\label{subsec:sso}
Subsequently, we tested SSO, i.e. the recently introduced replacement for the NSP objective, which proved to be ineffective. We compared models trained with three values of SSO weight $\alpha$ (see Section \ref{par:training_objective}): 0.0 (no SSO), 0.1 (small impact of SSO), and 1.0 (SSO equally important as MLM objective) (see Table \ref{tab:sso-study}).

The experiment showed, that SSO actually hurts downstream task performance. The differences between enabled and disabled SSO are statistically significant for two out of three experimental setups. The only scenario for which the negative effect of SSO is not statistically significant is using the Large corpus and BPE-dropout. Overall, the best results are achieved using a small SSO weight but the differences are not significantly different from disabling SSO.

\begin{table}[!ht]
\renewcommand*{\arraystretch}{1.15}
\centering
\begin{tabular}{l|llr}

    \toprule
    \bf{SSO} & \bf{Corpus} & \bf{BPE} & \bf{Score} \\
    \midrule
    0.0 & Small & Yes & \underline{89.18 $\pm$ 0.15} \\
    0.1 & Small & Yes & 88.90 $\pm$ 0.24 \\
    1.0 & Small & Yes & 88.73 $\pm$ 0.08 \\
    \midrule
    0.0 & Large & Yes & 89.25 $\pm$ 0.21 \\
    0.1 & Large & Yes & \underline{89.37 $\pm$ 0.25} \\
    1.0 & Large & Yes & 89.10 $\pm$ 0.19 \\
    \midrule
    0.0 & Large & No & 89.28 $\pm$ 0.26 \\
    0.1 & Large & No & \bf{\underline{89.45 $\pm$ 0.18}} \\
    1.0 & Large & No & 88.80 $\pm$ 0.15 \\
    \bottomrule

\end{tabular}
\caption{Average scores on KLEJ Benchmark depending on a SSO weight. We used BERT\textsubscript{BASE} model trained for 10k iterations with BPE-Dropout. The best score within each group is underlined, the best overall is bold.}
\label{tab:sso-study}
\end{table}

\paragraph{BPE-Dropout}
\label{subsec:dropoutbpe}
The BPE-Dropout could be beneficial for downstream task performance, but its impact is difficult to assess due to many confounding variables.

The model initialization with XLM-RoBERTa weights means that token embedding is already semantically meaningful even without additional pretraining. However, for both random and XLM-RoBERTa initialization the BPE-Dropout is beneficial.

% wariant z inicjalizacją
\begin{table}[!ht]
\renewcommand*{\arraystretch}{1.15}
\setlength{\tabcolsep}{4.5pt}
\centering
\begin{tabular}{l|lllr}

    \toprule
    \bf{BPE} & \bf{Init} & \bf{SSO} & \bf{Corpus} & \bf{Score} \\
    \midrule
    No & Random & 1.0 & Large & 85.65 $\pm$ 0.43 \\
    Yes & Random & 1.0 & Large & \underline{85.78 $\pm$ 0.23} \\
    \midrule
    No & XLM-R & 1.0 & Large & 88.80 $\pm$ 0.15 \\
    Yes & XLM-R & 1.0 & Large & \underline{89.10 $\pm$ 0.19} \\
    \midrule
    No & XLM-R & 0.0 & Large & \bf{\underline{89.28 $\pm$ 0.26}} \\
    Yes & XLM-R & 0.0 & Large & 89.25 $\pm$ 0.21 \\
    \midrule
    No & XLM-R & 0.0 & Small & 89.12 $\pm$ 0.29 \\
    Yes & XLM-R & 0.0 & Small & \underline{89.18 $\pm$ 0.15} \\
    \bottomrule

\end{tabular}
\caption{Average scores on KLEJ Benchmark depending on usage of BPE-Dropout. We used BERT\textsubscript{BASE} model trained for 10k iterations on a large corpus. The best score within each group is underlined, the best overall is bold.}
\label{tab:dropoutbpe-study}
\end{table}

According to the results (see Table \ref{tab:dropoutbpe-study}), none of the reported differences is statistically significant and we can only conclude that BPE-Dropout does not influence the model performance.

\paragraph{Length of Pretraining}
\label{subsec:length}
The length of pretraining in terms of the number of iterations is commonly considered an important factor of the final quality of the model \citep{kaplan2020scaling}. Even though it seems straightforward to validate this hypothesis in practice it is not so trivial.

When pretraining Transformer-based models linear decaying learning rate is typically used. Therefore, increasing the number of training iterations changes the learning rate schedule and impacts the training. In our initial experiments usage of the same learning rate caused the longer training to collapse. Instead, we chose the learning rate for which the value of loss function after 10k steps was similar. We found that the learning rate equal to $3 \cdot 10^{-4}$ worked best for training in 50k steps.

\begin{table}[!ht]
\renewcommand*{\arraystretch}{1.15}
\centering
\begin{tabular}{ll|lr}

    \toprule
    \bf{\# Iter} & \bf{LR} & \bf{SSO} & \bf{Score} \\
    \midrule
    10k & $7 \cdot 10^{-4}$ & 1.0 & 89.10 $\pm$ 0.19 \\
    50k & $3 \cdot 10^{-4}$ & 1.0 & \underline{89.43 $\pm$ 0.10} \\
    \midrule
    10k & $7 \cdot 10^{-4}$ & 0.1 & 89.37 $\pm$ 0.25 \\
    50k & $3 \cdot 10^{-4}$ & 0.1 & \bf{\underline{89.87 $\pm$ 0.22}} \\
    \bottomrule

\end{tabular}
\caption{Average scores on KLEJ Benchmark depending training length. We used BERT\textsubscript{BASE} model trained on a large corpus with BPE-Dropout. The best score within each group is underlined, the best overall is bold.}
\label{tab:length-study}
\end{table}

Using the presented experiment setup, we tested the impact of pretraining length for two values of SSO weight: 1.0 and 0.1. In both cases, the model pretrained with more iterations achieves only slightly better but statistically significant results (see Table \ref{tab:length-study}).

\section{HerBERT}
\label{sec:herbert}
In this section, we apply conclusions drawn from the ablation study (see Section \ref{sec:ablation}) and describe the final pretraining procedure used to train HerBERT model.

\paragraph{Pretraining Procedure}
HerBERT was trained on the \emph{Large} corpus. We used Dropout-BPE in tokenizer with a probability of a drop equals to 10\%. Finally, HerBERT models were initialized with weights from XLM-RoBERTa and were trained with the objective defined in Equation \ref{eq:loss} with SSO weight equal to 0.1.

\begin{table*}
\renewcommand*{\arraystretch}{1.15}
\centering
\begin{tabular}{l|c|ccccccccc}

    \toprule
    \bf{Model} & \rot{\bf{AVG }} & \rot{\bf{NKJP-NER }} & \rot{\bf{CDSC-E }} & \rot{\bf{CDSC-R }} & \rot{\bf{CBD }} & \rot{\bf{PolEmo2.0-IN }} & \rot{\bf{PolEmo2.0-OUT }} & \rot{\bf{Czy wiesz? }} & \rot{\bf{PSC }} & \rot{\bf{AR }} \\
    \toprule
    \multicolumn{11}{c}{\bf{Base Models}} \\
    \midrule
    XLM-RoBERTa    & 84.7 $\pm$ 0.29 & 91.7 & 93.3 & 93.4 & 66.4 & \underline{90.9} & 77.1 & 64.3 & 97.6 & 87.3 \\
    Polish RoBERTa & 85.6 $\pm$ 0.29 & 94.0 & 94.2 & \underline{94.2} & 63.6 & 90.3 & 76.9 & \underline{71.6} & 98.6 & 87.4 \\
    HerBERT        & \underline{86.3 $\pm$ 0.36} & \underline{94.5} & \underline{94.5} & 94.0 & \underline{67.4} & \underline{90.9} & \underline{80.4} & 68.1 & \underline{98.9} & \underline{87.7} \\
    \midrule
    \multicolumn{11}{c}{\bf{Large Models}} \\
    \midrule
    XLM-RoBERTa    & 86.8 $\pm$ 0.30 & 94.2 & \bf{\underline{94.7}} & 93.9 & 67.6 & 92.1 & 81.6 & 70.0 & 98.3 & 88.5 \\
    Polish RoBERTa & 87.5 $\pm$ 0.29 & 94.9 & 93.4 & 94.7 & 69.3 & \bf{\underline{92.2}} & 81.4 & 74.1 & \bf{\underline{99.1}} & 88.6 \\
    HerBERT        & \bf{\underline{88.4 $\pm$ 0.19}} & \bf{\underline{96.4}} & 94.1 & \bf{\underline{94.9}} & \bf{\underline{72.0}} & \bf{\underline{92.2}} & \bf{\underline{81.8}} & \bf{\underline{75.8}} & 98.9 & \bf{\underline{89.1}} \\
    \bottomrule

\end{tabular}
\caption{Evaluation results on KLEJ Benchmark. \textit{AVG} is the average score across all tasks. Scores are reported for test set and  correspond to median values across five runs. The best scores within each group are underlined, the best overall are in bold.}
\label{tab:results-klej}
\end{table*}

\paragraph{Optimization}
We trained HerBERT\textsubscript{BASE} using Adam optimizer \citep{Kingma2015Adam} with parameters: $\beta_{1} = 0.9$, $\beta_{2} = 0.999$, $\epsilon = 10^{-8}$ and a linear decay learning rate schedule with a peak value of $3 \cdot 10^{-4}$. Due to the initial transfer of weights from the already trained model, the warm-up stage was set to a relatively small number of 500 iterations. The whole training took 50k iterations. 

Training of HerBERT\textsubscript{LARGE} was longer (60k iterations) and had a more complex learning rate schedule. For the first 15k we linearly decayed the learning rate from $3 \cdot 10^{-4}$ to $2.5 \cdot 10^{-4}$. We observed the saturation of evaluation metrics and decided to drop the learning rate to $1 \cdot 10^{-4}$. After training for another 25k steps and reaching the learning rate of $7 \cdot 10^{-5}$ we again reached the plateau of evaluation metrics. In the last phase of training, we dropped the learning rate to $3 \cdot 10^{-5}$ and trained for 20k steps until it reached zero. Additionally, during the last phase of training, we disabled both BPE-Dropout and dropout within the Transformer itself as suggested by \citet{albert2020iclr}.

Both HerBERT\textsubscript{BASE} and HerBERT\textsubscript{LARGE} models were trained with a batch size of 2560.

\section{Evaluation}
\label{sec:eval}
In this section, we introduce other top-performing models for Polish language understanding and compare their performance on evaluation tasks (see Section \ref{subsec:tasks}) to HerBERT.

\subsection{Models}
\label{subsec:models}
According to the KLEJ Benchmark leaderboard\footnote{\url{https://klejbenchmark.com/leaderboard/}} the three top-performing models of Polish language understanding are XLM-RoBERTa-NKJP\footnote{XLM-RoBERTa-NKJP is XLM-RoBERTa model additionally fine-tuned on NKJP corpus.}, Polish RoBERTa, and XLM-RoBERTa. These are also the only three models available in LARGE architecture variant.

Unfortunately, the XLM-RoBERTa-NKJP model is not publicly available, so we cannot use it for our evaluation. However, it has the same average score as the runner-up (Polish RoBERTa) which we compare HerBERT with.

\subsection{Results}
\paragraph{KLEJ Benchmark}

Both variants of HerBERT achieved the best average performance, significantly outperforming Polish RoBERTa and XLM-RoBERTa (see Table \ref{tab:results-klej}). Regarding BASE models, HerBERT\textsubscript{BASE} improves the state-of-the-art result by 0.7pp and for HerBERT\textsubscript{LARGE} the improvement is even bigger (0.9pp). In particular, HerBERT\textsubscript{BASE} scores best on eight out of nine tasks (tying on PolEmo2.0-OUT, performing slightly worse on CDSC-R) and HerBERT\textsubscript{LARGE} in seven out of nine tasks (tying in PolEmo2.0-IN, performing worse in CDSC-E and PSC). Surprisingly, HerBERT\textsubscript{BASE} is better than HerBERT\textsubscript{LARGE} in CDSC-E. The same behaviour is noticeable for Polish RoBERTa, but not for XLM-RoBERTa. For other tasks, the LARGE models are always better. Summing up, HerBERT\textsubscript{LARGE} is the new state-of-the-art Polish language model based on the results of the KLEJ Benchmark. 

It is worth emphasizing that the proposed procedure for efficient pretraining by transferring knowledge from multilingual to monolingual language allowed HerBERT to achieve better results than Polish RoBERTa even though it was optimized with around ten times shorter training.

\paragraph{POS Tagging}

\begin{table}[!ht]
\renewcommand*{\arraystretch}{1.15}
\centering
\begin{tabular}{lrr}

    \toprule
    \bf{Model} & \bf{Accuracy } & \bf{F1-Score } \\
    \midrule
    \multicolumn{3}{c}{\bf{Base Models}} \\
    \midrule
    XLM-RoBERTa    & 95.97 $\pm$ 0.04 & 95.79 $\pm$ 0.05 \\
    Polish RoBERTa & 96.13 $\pm$ 0.03 & 95.92 $\pm$ 0.03 \\
    HerBERT        & \underline{96.46 $\pm$ 0.04} & \underline{96.27 $\pm$ 0.04} \\
    \midrule
    \multicolumn{3}{c}{\bf{Large Models}} \\
    \midrule
    XLM-RoBERTa    & 97.07 $\pm$ 0.05 & 96.93 $\pm$ 0.05 \\
    Polish RoBERTa & 97.21 $\pm$ 0.02 & 97.05 $\pm$ 0.03 \\
    HerBERT        & \bf{\underline{97.30 $\pm$ 0.02}} & \bf{\underline{97.17 $\pm$ 0.02}} \\
    \bottomrule

\end{tabular}
\caption{Part-of-speech tagging results on NKJP dataset. Scores are reported for the test set and are median values across five runs. Best scores within each group are underlined, best overall are bold.}
\label{tab:results-pos}
\end{table}

HerBERT achieves overall better results in terms of both accuracy and F1-Score. HerBERT\textsubscript{BASE} beats the second-best model (i.e. Polish RoBERTa) by a margin of 0.33pp (F1-Score by 0.35pp) and HerBERT\textsubscript{LARGE} by a margin of 0.09pp (F1-Score by 0.12pp). It should be emphasized that while the improvements may appear to be minor, they are statistically significant. 
All results are presented in Table \ref{tab:results-pos}.

\paragraph{Dependency Parsing}
The dependency parsing results are much more ambiguous than in other tasks. As expected, the models with static fastText embeddings performed much worse than Transformer-based models (around 3pp difference for UAS, and 4pp for LAS).
% Moreover, model initialized with fastText embeddings is better than the model without them. 
In the case of Transformer-based models, the differences are less noticeable. As expected, the LARGE models outperform the BASE models. The best performing model is Polish RoBERTa. HerBERT models performance is the worst across Transformer-based models except for the UAS score which is slightly better than XLM-RoBERTa for BASE models. 
All results are presented in Table \ref{tab:results-dep}.

\begin{table}[!ht]
\renewcommand*{\arraystretch}{1.15}
\setlength{\tabcolsep}{5pt}
\centering
\begin{tabular}{lrr}

    \toprule
    \bf{Model} & \bf{UAS } & \bf{LAS } \\
    \midrule
    \multicolumn{3}{c}{\bf{Static Embeddings}} \\
    \midrule
    Plain & 90.58 $\pm$ 0.07 & 87.35 $\pm$ 0.12 \\
    FastText      & \underline{92.20 $\pm$ 0.14} & \underline{89.57 $\pm$ 0.13} \\
    \midrule
    \multicolumn{3}{c}{\bf{Base Models}} \\
    \midrule
    XLM-RoBERTa    & 95.14 $\pm$ 0.07 & 93.25 $\pm$ 0.12 \\
    Polish RoBERTa & \underline{95.41 $\pm$ 0.24} & \underline{93.65 $\pm$ 0.34} \\
    HerBERT        & 95.18 $\pm$ 0.22 & 93.24 $\pm$ 0.23 \\
    \midrule
    \multicolumn{3}{c}{\bf{Large Models}} \\
    \midrule
    XLM-RoBERTa    & 95.38 $\pm$ 0.02 & 93.66 $\pm$ 0.07 \\
    Polish RoBERTa & \bf{\underline{95.60 $\pm$ 0.18}} & \bf{\underline{93.90 $\pm$ 0.21}} \\
    HerBERT        & 95.11 $\pm$ 0.04 & 93.32 $\pm$ 0.02 \\
    \bottomrule

\end{tabular}
\caption{Dependency parsing results on Polish Dependency Bank dataset. Scores are reported for the test set and are median values across three runs. Best scores within each group are underlined, best overall are bold.}
\label{tab:results-dep}
\end{table}

\section{Conclusion}
\label{sec:conclusion}
In this work, we conducted a thorough ablation study regarding training BERT-based models for Polish language. We evaluated several design choices for pretraining BERT outside of English language. Contrary to \citet{structbert2020iclr}, our experiments demonstrated that SSO is not beneficial for the downstream task performance. It also turned out that BPE-Dropout does not increase the quality of a pretrained language model.

As a result of our studies we developed and evaluated an efficient pretraining procedure for transferring knowledge from multilingual to monolingual BERT-based models. We used it to train and release HerBERT -- a Transformer-based language model for Polish. It was trained on a diverse multi-source corpus. The conducted experiments confirmed its high performance on a set of eleven diverse linguistic tasks, as HerBERT turned out to be the best on eight of them. In particular, it is the best model for Polish language understanding according to the KLEJ Benchmark.

It is worth emphasizing that the quality of the obtained language model was even more impressive considering its short training time. Due to multilingual initialization, HerBERT\textsubscript{BASE} outperformed Polish RoBERTa\textsubscript{BASE} even though it was trained with a smaller batch size (2560 vs 8000) for a fewer number of steps (50k vs 125k). The same behaviour is also visible for HerBERT\textsubscript{LARGE}.
Additionally, we conducted a separate ablation study to confirm that the success of HerBERT is caused by the described initialization scheme. It showed that in fact, it was the most important factor to improved the quality of HerBERT.

We believe that the proposed training procedure and detailed experiments will encourage NLP researchers to cost-effectively train language models for other languages.

\newpage

\bibliography{anthology,eacl2021}
\bibliographystyle{acl_natbib}

\end{document}